\documentclass[runningheads]{llncs}
\usepackage{graphicx}

\usepackage{float}
\usepackage{graphicx}
\usepackage{comment}
\usepackage{amsmath,amssymb} 
\usepackage{color}
\usepackage{xspace}
\usepackage{textcomp}
\usepackage{cite}
\usepackage{mathtools}
\usepackage{bbm}
\usepackage{makecell, verbatim, sidecap}
\usepackage{subfig}
\usepackage{multirow}
\usepackage{url}

\renewcommand{\vec}[1]{\ensuremath{\pmb{#1}}}
\newcommand{\mat}[1]{\ensuremath{\mathbf{#1}}}
\newcommand{\set}[1]{\ensuremath{\mathscr{#1}}}

\makeatletter
\@tfor\next:=abcdefghijklmnopqrstuvwxyxz\do
{\begingroup\edef\x{\endgroup
		\noexpand\@namedef{v\next}{\noexpand\vec{\next}}%
	}\x}
\@tfor\next:=ABCDEFGHIJKLMNOPQRSTUVWXYZ\do
{\begingroup\edef\x{\endgroup
		\noexpand\@namedef{m\next}{\noexpand\mat{\next}}%
	}\x}
\@tfor\next:=ABCDEFGHIJKLMNOPQRSTUVWXYZ\do
{\begingroup\edef\x{\endgroup
		\noexpand\@namedef{s\next}{\noexpand\set{\next}}%
	}\x}
\makeatother

\def\eg{{\it e.g.}\xspace}
\def\ie{{\it i.e.}\xspace}

\def\R{{\mathbb R}}

\def\pos{{\rm pos}}

\def\sigmoid{{\rm sigmoid}}
\def\ReLU{{\rm ReLU}}

\DeclarePairedDelimiter{\floor}{\lfloor}{\rfloor}

\usepackage[colorlinks,citecolor=blue,urlcolor=blue,bookmarks=false,hypertexnames=true]{hyperref}

 \usepackage[margin=8pt,font=small,labelfont={sc},labelsep=endash,tableposition=top]{caption}

\begin{document}
\pagestyle{headings}
\mainmatter

\def\Ours{{CondInst}}

\title{Conditional Convolutions for Instance Segmentation}


\titlerunning{Conditional Convolutions for Instance Segmentation}
%
\author{Zhi Tian,
~~ Chunhua Shen%
\thanks{Corresponding author.
},
~~		Hao Chen}

\authorrunning{Z. Tian et al.; Appearing in Proc.\ European Conf.\ Comp.\ Vis.\ (ECCV) 2020}
%
\institute{The University of Adelaide, Australia}
\maketitle

\begin{abstract}
\renewcommand\UrlFont{\color{blue}}
We propose a simple yet effective instance segmentation fra\-me\-wor\-k, termed CondInst (conditional convolutions for instance segmentation). Top-performing instance segmentation methods such as Mask R-CNN rely on ROI operations (typically ROIPool or ROIAlign) to obtain the final instance masks. In contrast, we propose to solve instance segmentation from a new perspective. Instead of using instance-wise ROIs as inputs to a network of fixed weights, we employ dynamic instance-aware networks, conditioned on instances. CondInst enjoys two advantages: 1) Instance segmentation is solved by a fully convolutional network, eliminating the need for ROI cropping and feature alignment. 2) Due to the much improved capacity of dynamically-generated conditional convolutions, the mask head can be very compact (\eg, 3 conv.\ layers, each having only 8 channels), leading to significantly faster inference. We demonstrate a simpler instance segmentation method that can achieve improved performance in both accuracy and inference speed. On the COCO dataset, we outperform a few recent methods including well-tuned Mask R-CNN baselines, without longer training schedules needed. Code is available: \texttt{\url{https://git.io/AdelaiDet}}

\end{abstract}

\begin{figure}[t]
\centering
\includegraphics[width=.68\textwidth]{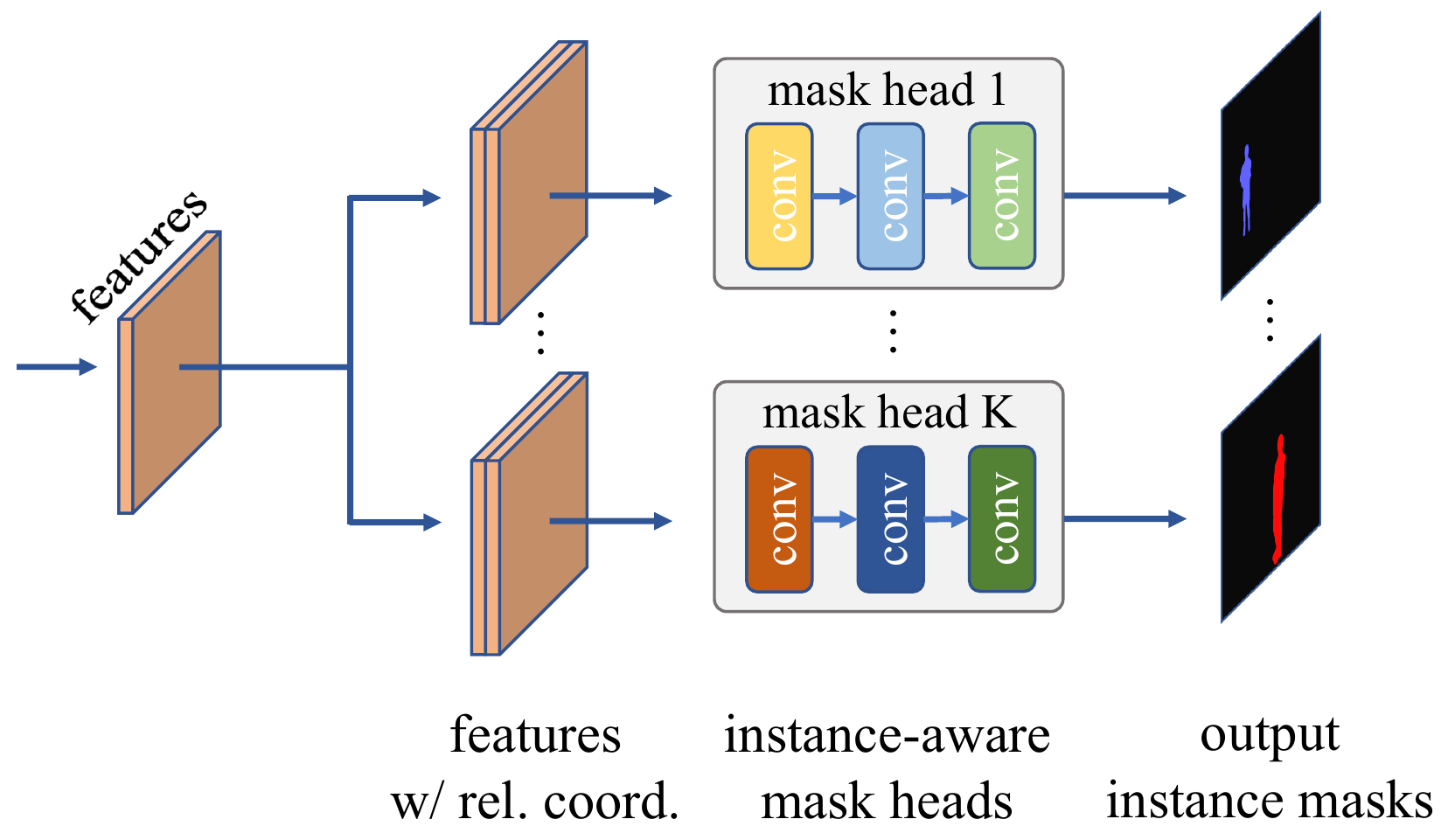}
\caption{\Ours\ uses instance-aware mask heads to predict the masks for each instance. $K$ is the number of instances to be predicted. The filters in the mask head vary with different instances, which are dynamically-generated and conditioned on the target instance. \ReLU\ is used as the activation function (excluding the last conv.\ layer).}
\label{fig:mask_heads}
\end{figure}

\section{Introduction}
Instance segmentation is a fundamental yet challenging task in computer vision, which requires
an
algorithm to predict a per-pixel mask with a category label for each instance of interest in an image.
Despite a few works being proposed recently,
the dominant framework for instance segmentation is still the two-stage
method Mask R-CNN \cite{he2017mask}, which casts instance segmentation into a two-stage detection-and-segmentation task. Mask R-CNN first employs an object detector Faster R-CNN
to predict a bounding-box  for  each instance.
Then for each instance, regions-of-interest (ROIs) are cropped from the networks' feature maps using the ROIAlign operation.
To predict the final masks for each instance, a compact
fully convolutional network (FCN) (\ie, mask head) is applied to these ROIs
to perform foreground/background segmentation.
However, this ROI-based method may have
the following drawbacks. 1) Since
ROIs are often axis-aligned bounding-boxes, for
objects with irregular shapes, they may contain an excessive amount of irrelevant image content including background and other instances.
This issue may be
mitigated by
using
rotated ROIs,
but with the price of a more complex pipeline.
2) In order to distinguish between the foreground instance and the background stuff or instance(s), the mask head requires a relatively larger receptive field to encode
sufficiently
large context information.
As a result, a stack of $3 \times 3$ convolutions is needed in the mask head (\eg, four $3 \times 3$ convolutions with $256$ channels in Mask R-CNN).
It
considerably
increases
computational complexity
of the mask head, resulting that the inference time significantly varies
in
the number of instances.
3)
ROIs are typically
of
different sizes. In order to use effective batched computation in modern deep learning frameworks \cite{pytorch, tensorflow}, a resizing operation is often required to resize the cropped regions into patches
of the same size.
For instance, Mask R-CNN resizes all the cropped regions to $14 \times 14$ (upsampled to $28 \times 28$ using a deconvolution), which
restricts
the output resolution of instance segmentation, as
large instances would require
higher resolutions to retain details at the boundary.

In computer vision,
the
closest
task to instance segmentation is semantic segmentation, for which fully convolutional networks (FCNs) have shown dramatic success \cite{long2015fully, chen2017deeplab, tian2019decoders, he2019knowledge, liu2020structured}. FCNs also have shown excellent performance on many other
per-pixel prediction
tasks
ranging from low-level image processing such as denoising, super-resolution;
to mid-level tasks such as optical flow estimation and contour detection;
and high-level tasks including
recent single-shot object detection \cite{tian2019fcos}, monocular depth estimation \cite{Depth2015Liu, Yin2019enforcing, yin2020diversedepth, bian2019unsupervised, bian2020unsupervised} and counting \cite{boominathan2016crowdnet}.
However,
almost all the instance segmentation methods based on FCNs\footnote{By FCNs, we mean the
	vanilla
	FCNs in \cite{long2015fully} that
	only
	involve convolutions and pooling.
} lag behind state-of-the-art ROI-based methods.
{Why do the versatile FCNs perform unsatisfactorily on instance segmentation?}
We observe that the major difficulty of applying FCNs to instance segmentation is that the similar image appearance may require different predictions but FCNs struggle at achieving this. For example, if two persons A and B with the similar appearance are in an input image, when predicting the instance mask of A,
the FCN needs to predict
B
as background w.r.t.\
A, which can be difficult as they
look similar in
appearance. Therefore, an ROI operation is %
used
to crop the person of interest, \ie, A; and
filter out B. Essentially, instance segmentation needs two types of information:
1) \textit{appearance} information to categorize objects; and
2) \textit{location} information to distinguish
multiple objects belonging to the same category.
Almost all methods rely on ROI cropping,
which explicitly encodes the location information of  instances.
In contrast,
\Ours\ exploits the location information by
using \textit{instance-sensitive convolution filters } as well as relative coordinates that are appended to the feature maps.

\begin{figure*}[t]
	\centering
	\includegraphics[width=.9\linewidth]{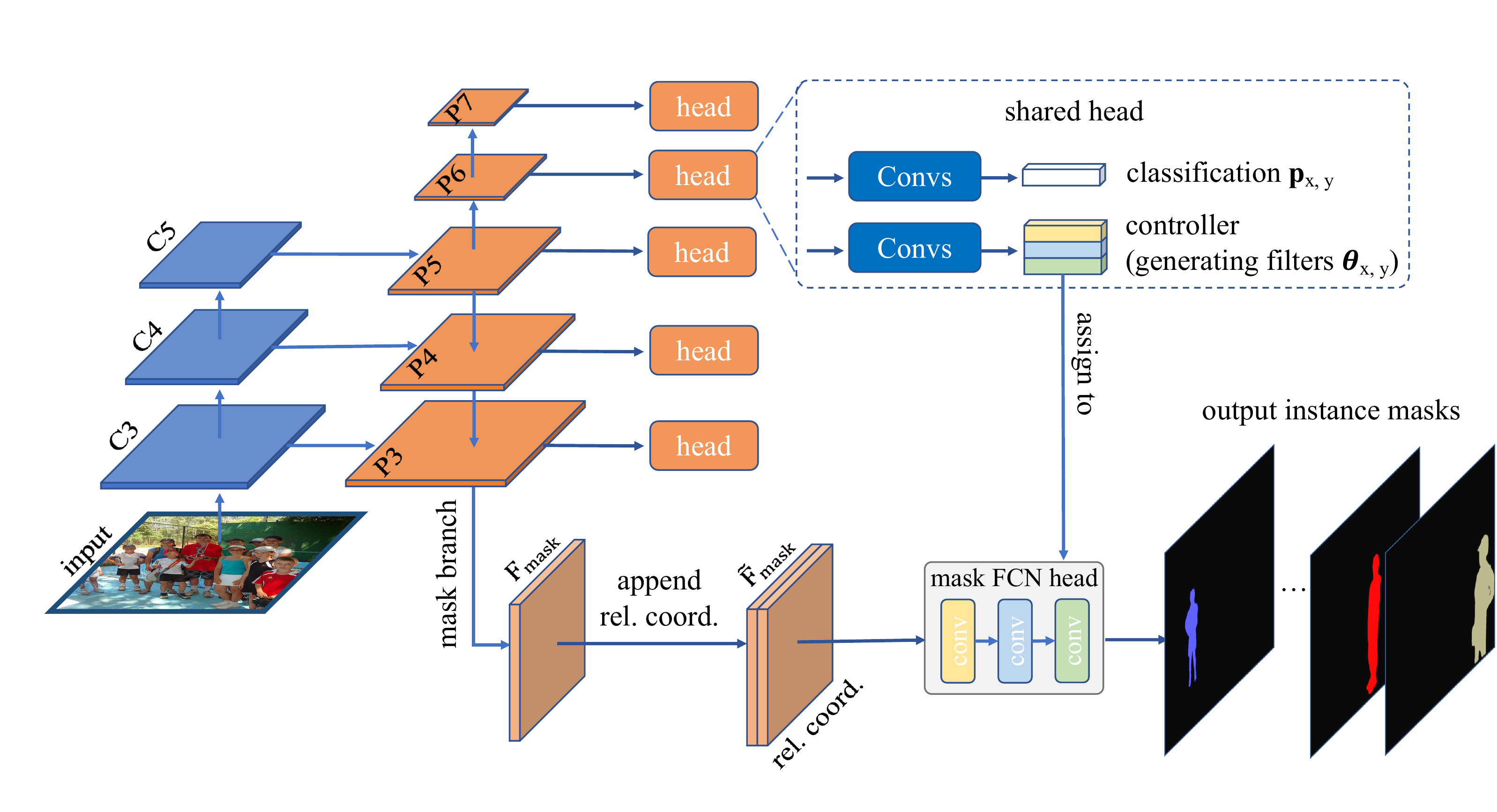}
	\caption{The overall architecture of \Ours. $C_3$, $C_4$ and $C_5$ are the feature maps of the backbone network (\eg, ResNet-50). $P_3$ to $P_7$ are the FPN feature maps as in \cite{lin2017feature, tian2019fcos}. $\mF_{mask}$ is the mask branch's output and $\tilde{\mF}_{mask}$ is obtained by concatenating  the relative coordinates to $\mF_{mask}$. The classification head predicts the class probability $\vp_{x, y}$ of the target instance at location $(x, y)$, same as in FCOS. The controller generates the filter parameters $\vec{\theta}_{x, y}$ of the mask head for the instance. Similar to FCOS, there are also center-ness and box heads in parallel with the controller (not shown in the figure for simplicity). Note that the heads in the dashed box are repeatedly applied to  $P_3$ $\cdots$$P_7$. The mask head is instance-aware, and is applied to $\tilde{\mF}_{mask}$ as many times as the number of instances in the image (refer to Fig.~\ref{fig:mask_heads}).}
	\label{fig:main_figure}
\end{figure*}

Thus,
we
advocate  a new
solution that uses
instance-aware
FCNs %
for instance mask prediction.
In other words, instead of
using
a standard ConvNet
with a fixed set of convolutional filters
as
the mask head
for predicting all instances,
the network parameters are
adapted
according to the instance to be predicted. Inspired by dynamic filtering networks \cite{jia2016dynamic} and CondConv \cite{yang2019condconv}, for each instance,
a controller sub-network (see Fig.~\ref{fig:main_figure})
dynamically generates the mask FCN network  parameters
(conditioned on the center area of the instance), which is then used to predict the mask of this instance. It is expected that the network parameters can encode the characteristics (\eg, relative position, shape and appearance) of this instance, and only fires on the pixels of this instance, which thus bypasses the difficulty mentioned above. These conditional mask heads are applied to the whole feature maps, \textit{eliminating the need for ROI operations}. At the first glance, the idea may
not work well
as instance-wise mask heads
may
incur a large number of network parameters provided that some images contain as many as dozens of instances. However,
we show
that a
very
compact FCN %
mask head
with dynamically-generated filters can already outperform previous ROI-based Mask R-CNN, resulting in much reduced computational complexity per instance than
that of the mask head in Mask R-CNN.

We summarize our main contributions as follow.
\begin{itemize}
	\itemsep 0cm
	\item
	We attempt to solve instance segmentation
	from a new perspective.
	To this end, we
	propose
	the \Ours\
	instance segmentation framework,
	which achieves improved instance segmentation performance than
	existing
	methods such as Mask R-CNN while being faster. To our knowledge, this is the first time that a
	new
	instance segmentation framework outperforms %
	recent state-of-the-art
	both in accuracy and speed.

	\item
	\Ours\
	is
	fully convolutional and avoids the aforementioned resizing operation used in %
	many existing
	methods, as \Ours\ does not rely on ROI operations.
	Without having to resize feature maps leads to high-resolution instance masks with more accurate edges.

	\item Unlike previous
	methods, in which the filters in its mask head are fixed for all the instances once trained, the filters in our mask head are dynamically generated and conditioned on instances.
	As the
	filters are only
	asked
	to predict the mask of only one instance,
	it
	largely eases the learning requirement and thus reduces the load of the filters. As a result, the mask head can be extremely light-weight, significantly reducing the inference time per instance.
	Compared with the bounding box detector FCOS,
	\Ours\ needs only  $\sim$10\% more computational time, even processing the maximum number of instances per image (\ie, $100$ instances).

	\item Without resorting to longer training schedules as needed  in recent works \cite{chen2019tensormask, bolya2019yolact}, \Ours\ achieves state-of-the-art performance while being faster in inference. We hope that \Ours\ can be a new strong alternative to popular methods such as Mask R-CNN for the instance segmentation task.

\end{itemize}

Moreover, \Ours\
can
be immediately applied
to panoptic segmentation due to its flexible design.
We believe that with minimal re-design effort,
the proposed \Ours\ can be used to solve all instance-level recognition tasks that were previously solved with an ROI-based pipeline.

\begin{figure*}[t]
	\centering
	\includegraphics[width=1.09\linewidth]{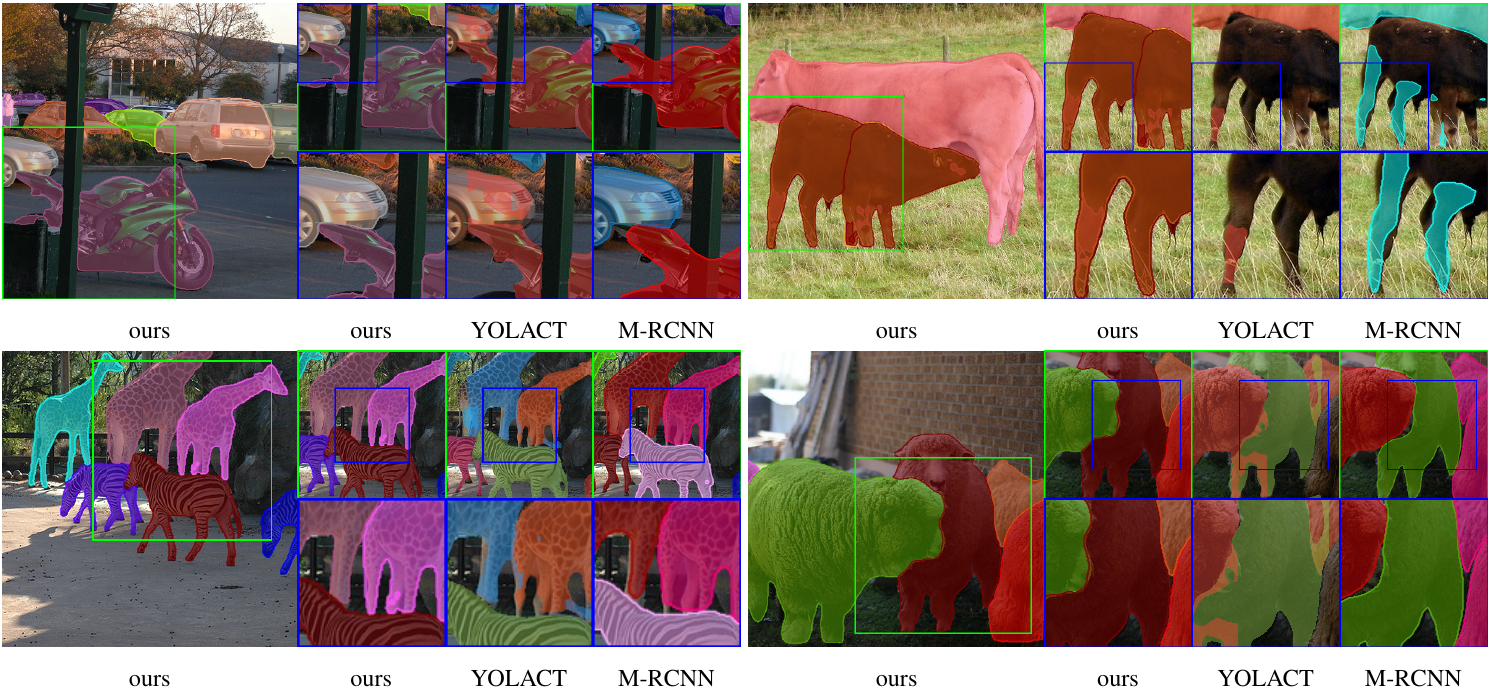}
	\caption{Qualitative comparisons with other methods. We compare the proposed \Ours\ against YOLACT~\cite{bolya2019yolact} and Mask R-CNN~\cite{he2017mask}. Our masks are generally of higher quality (\eg, preserving more details). Best viewed on screen.}
	\label{fig:qualitative}
\end{figure*}

\subsection{Related Work}
Here we review some work that is most relevant to ours.

\noindent\textbf{Conditional Convolutions.} Unlike traditional convolutional layers, which have fixed filters once trained, the filters of conditional convolutions are conditioned on the input and are dynamically generated by another network (\ie, a controller). This idea has been
explored previously
in dynamic filter networks \cite{jia2016dynamic} and CondConv \cite{yang2019condconv} mainly for the purpose
of
increasing  the capacity of a classification network. In this work, we
extend
this idea to solve the significantly more challenging task of
instance segmentation.

\noindent\textbf{Instance Segmentation.}
To date,
the dominant framework for  instance segmentation is still Mask R-CNN. Mask R-CNN first employs an object detector to detect the bounding-boxes of instances (\eg, ROIs). With these bounding-boxes, an ROI operation is used to crop the features of the instance from the feature maps. Finally, a
compact
FCN head is
used
to obtain the desired instance masks. Many works \cite{chen2019hybrid, liu2018path, huang2019mask} with top performance are built on Mask R-CNN.
Moreover, some works have explored to apply FCNs to instance segmentation.
InstanceFCN \cite{dai2016instance} may be the first instance segmentation method
that
is fully convolutional. InstanceFCN proposes to predict position-sensitive score maps with vanilla FCNs. Afterwards, these score maps are assembled to obtain the desired instance masks.
Note that
InstanceFCN does not work well with overlapping instances.
Others
\cite{neven2019instance, newell2017associative, fathi2017semantic} attempt to first
perform
segmentation and the desired instance masks are formed by assembling the pixels of the same instance. Novotny et al.\ \cite{Novotny_2018_ECCV} propose semi-convolutional operators to make FCNs applicable to instance segmentation. To our knowledge, thus far
none of
these
methods can outperform Mask R-CNN both in accuracy and speed on
the
public
COCO benchmark dataset.

The recent YOLACT \cite{bolya2019yolact}
and BlendMask \cite{chen2020blendmask} may be viewed as a reformulation of Mask RCNN, which decouple ROI detection and
feature maps used for mask prediction.
Wang et al.\ developed a simple FCN based instance segmentation method, showing competitive performance \cite{wang2019solo}. PolarMask developed a new simple mask
representation for instance segmentation \cite{polarmask},
which  extends  the bounding box detector FCOS \cite{tian2019fcos}. EmbedMask~\cite{ying2019embedmask} learns instance and pixel embedding, and then assigns pixels to an instance based on the similarity of their embedding.

Recently AdaptIS \cite{sofiiuk2019adaptis} proposes to solve panoptic segmentation with FiLM \cite{perez2018film}. The idea
shares some similarity  with \Ours\
in that information about an instance is encoded in the coefficients generated by FiLM.
Since only the batch normalization coefficients are dynamically generated, AdaptIS
needs a large mask head to achieve good performance. In contrast, \Ours\ directly encodes them into conv.\  filters of the mask head, thus having much stronger capacity.
As a result,
even with a very compact mask head, we believe that
\Ours\ can achieve instance segmentation accuracy that would not be possible for AdaptIS to attain.

\section{Instance Segmentation with \Ours}

\subsection{Overall Architecture}
Given an input image $I \in \R^{H \times W \times 3}$, the goal of instance segmentation is to predict the pixel-level mask and the category of each instance of interest in the image. The ground-truths are defined as $\{(M_i, c_i)\}$, where $M_i \in \{0, 1\}^{H \times W}$ is the mask for the $i$-th instance and $c_i \in \{1, 2, ..., C\}$ is the category. $C$ is $80$ on MS-COCO \cite{lin2014microsoft}. Unlike semantic segmentation, which only requires to predict one mask for an input image, instance segmentation needs to predict a variable number of masks, depending on the number of instances in the image. This poses a challenge when applying traditional FCNs~\cite{long2015fully} to instance segmentation. In this work, our core idea is that for an image with $K$ instances, $K$ different mask heads will be dynamically generated, and each mask head will contain the characteristics of its target instance in their filters. As a result, when the mask is applied to an input, it will only fire on the pixels of the instance, thus producing the mask prediction of the instance. We illustrate the process in Fig.~\ref{fig:mask_heads}.

Recall that
Mask R-CNN employs an object detector to predict the bounding-boxes of the instances in the input image. The bounding-boxes are actually the way that Mask R-CNN represents instances. Similarly, \Ours\ employs the instance-aware filters to represent the instances. In other words, instead of encoding the instance concept into the bounding-boxes, \Ours\ implicitly encodes it into the parameters of the mask heads, which is a much more flexible way. For example, it can easily represent the irregular shapes that are hard to be tightly enclosed by a bounding-box. This is one of \Ours's advantages over the previous ROI-based methods.

Similar to the way that ROI-based methods obtain bounding-boxes, the instance-aware filters can also be obtained with an object detector. In this work, we build \Ours\ on the popular object detector FCOS~\cite{tian2019fcos} due to its simplicity and flexibility. Also, the elimination of anchor-boxes in FCOS can also save the number of parameters and the amount of computation of \Ours. As shown in Fig.~\ref{fig:main_figure}, following FCOS \cite{tian2019fcos}, we make use of the feature maps $\{P_3, P_4, P_5, P_6, P_7\}$ of feature pyramid networks (FPNs) \cite{lin2017feature}, whose down-sampling ratios are $8$, $16$, $32$, $64$ and $128$, respectively. As shown in Fig.~\ref{fig:main_figure}, on each feature level of the FPN, some functional layers (in the dash box) are applied to make instance-related predictions. For example, the class of the target instance and the dynamically-generated filters for the instance. In this sense, \Ours\ can be viewed as the same as Mask R-CNN, both of which first attend to instances in an image and then predict the pixel-level masks of the instances (\ie, instance-first).

Besides the detector, as shown in Fig.~\ref{fig:main_figure}, there is also a mask branch, which provides the feature maps that our generated mask heads take as inputs to predict the desired instance mask. The feature maps are denoted by \[
\mF_{mask} \in \R^{H_{mask} \times W_{mask} \times C_{mask}}.
\]
 The mask branch is connected to FPN level $P_3$ and thus its output resolution is $\frac{1}{8}$ of the input image resolution. The mask branch has four $3 \times 3$ convolutions with $128$ channels before the last layer. Afterwards, in order to reduce the number of the generated parameters, the last layer of the mask branch reduces the number of channels from $128$ to $8$ (\ie, $C_{mask} = 8$). Surprisingly, using $C_{mask} = 8$ can already achieve superior performance and using a larger $C_{mask}$ here (\eg, 16) cannot improve the performance, as shown in our experiments. Even more aggressively, using $C_{mask} = 2$ only degrades the performance by $\sim0.3\%$ in mask AP. Moreover, as shown in Fig.~\ref{fig:main_figure}, $\mF_{mask}$ is combined with a map of the coordinates, which are relative coordinates from all the locations on $\mF_{mask}$ to the location $(x, y)$ (\ie, where the filters of the mask head are generated). Then, the combination is sent to the mask head to predict the instance mask. The relative coordinates provide a strong cue for predicting the instance mask, as shown in our experiments. Moreover, a single \sigmoid\ is used as the final output of the mask head, and thus the mask prediction is class-agnostic. The class of the instance is predicted by the classification head in parallel with the controller, as shown in Fig.~\ref{fig:main_figure}.

The resolution of the original mask prediction is same as the resolution of $F_{mask}$, which is $\frac{1}{8}$ of the input image resolution. In order to produce high-resolution instance masks, a bilinear upsampling is used to upsample the mask prediction by $4$, resulting in $400 \times 512$ mask prediction (if the input image size is $800 \times 1024$). We will show that the upsampling is crucial to the final instance segmentation performance of \Ours\ in experiments. Note that the mask's resolution is much higher than that of Mask R-CNN (only $28 \times 28$ as mentioned before).

\subsection{Network Outputs and Training Targets}
Similar to FCOS, each location on the FPN's feature maps ${P_i}$ either is associated with an instance, thus being a positive sample, or is considered a negative sample. The associated instance and label for each location are determined as follows. Let us consider the feature maps $P_i \in \R^{H \times W \times C}$ and let $s$ be its down-sampling ratio. As shown in previous works \cite{tian2019fcos, ren2015faster, he2015spatial}, a location $(x, y)$ on the feature maps can be mapped back onto the input image as $(\floor{\frac{s}{2}} + xs, \floor{\frac{s}{2}} + ys)$. If the mapped location falls in the center region of an instance, the location is considered to be responsible for the instance. Any locations outside the center regions are labeled as negative samples. The center region is defined as the box $(c_x - rs, c_y - rs, c_x + rs, c_y + rs)$, where $(c_x, c_y)$ denotes the mass center of the instance, $s$ is the down-sampling ratio of $P_i$ and $r$ is a constant scalar being $1.5$ as in FCOS \cite{tian2019fcos}. As shown in Fig.~\ref{fig:main_figure}, at a location $(x, y)$ on $P_i$, \Ours\ has the following output heads.

\paragraph{Classification Head.} The classification head predicts the class of the instance associated with the location. The ground-truth target is the instance's class $c_i$ or $0$ (\ie, background). As in FCOS, the network predicts a $C$-D vector $\vp_{x, y}$ for the classification and each element in $\vp_{x, y}$ corresponds to a binary classifier, where $C$ is the number of categories.

\paragraph{Controller Head.} The controller head, which has the same architecture as the above classification head, is used to predict the parameters of the mask head for the instance at the location. The mask head predicts the mask of this particular instance. This is the core contribution of our work. To predict the parameters, we concatenate all the parameters of the filters (\ie, weights and biases) together as an $N$-D vector $\vec{\theta}_{x, y}$, where $N$ is the total number of the parameters. Accordingly, the controller head has $N$ output channels. The mask head is a very compact FCN architecture, which has three $1 \times 1$ convolutions, each having $8$ channels and using \ReLU\ as the activation function except for the last one. No normalization layer such as batch normalization \cite{ioffe2015batch} is used here. The last layer has $1$ output channel and uses \sigmoid\ to predict the probability of being foreground. The mask head has $169$ parameters in total ($\#weights = (8 + 2) \times 8 (conv1) + 8 \times 8 (conv2) + 8 \times 1 (conv3)$ and $\#biases = 8 (conv1) + 8 (conv2) + 1 (conv3)$). As mentioned before, the generated filters contain information about the instance at the location, and thus, ideally, the mask head with the filters will only fire on the pixels of the instance, even taking as the input the whole feature maps.

\paragraph{Center-ness and Box Heads.} The center-ness and box heads are the same as that in FCOS. We refer readers to FCOS \cite{tian2019fcos} for the details.

Conceptually, \Ours\ can eliminate the box head since \Ours\ needs no ROIs. However, we find that if we make use of box-based NMS, the inference time will be much reduced. Thus, we still predict boxes in \Ours. We would like to highlight that the predicted boxes are \emph{only} used in NMS and do not involve any ROI operations. Moreover, as shown in Table~\ref{table:mask_nms}, the box prediction can be removed if the NMS using no box (\eg, mask NMS or peak NMS \cite{zhou2019objects}) used. This is fundamentally different from previous ROI-based methods, in which the box prediction is mandatory.

\subsection{Loss Function}
Formally, the overall loss function of \Ours\ can be formulated as,
\begin{equation} \label{loss_function_overall}
\begin{aligned}
L_{overall} = L_{fcos} + \lambda L_{mask},
\end{aligned}
\end{equation}
where $L_{fcos}$ and $L_{mask}$ denote the original loss of FCOS and the loss for instance masks, respectively. $\lambda$ being $1$ in this work is used to balance the two losses. We refer readers to FCOS for the details of $L_{fcos}$. $L_{mask}$ is defined as,
\begin{equation} \label{loss_function_mask}
\begin{aligned}
&
L_{mask}(\{\vec{\theta}_{x, y}\})  = \frac{1}{N_{\pos}}\sum_{{x, y}}{\mathbbm{1}_{\{c^*_{x, y} > 0\}}L_{dice}(MaskHead(\mat{\tilde{F}}_{x, y}; \vec{\theta}_{x, y}), \mM^*_{x, y})},
\end{aligned}
\end{equation}
where $c^*_{x, y}$ is the classification label of location $(x, y)$, which is the class of the instance associated with the location or $0$ (\ie, background) if the location is not associated with any instance. $N_{pos}$ is the number of locations where $c^*_{x, y} > 0$. $\mathbbm{1}_{\{c^*_{x, y} > 0\}}$ is the indicator function, being $1$ if $c^*_{x, y} > 0$ and $0$ otherwise. $\vec{\theta}_{x, y}$ is the generated filters' parameters at location $(x, y)$. $\mat{\tilde{F}}_{x, y} \in \R^{H_{mask} \times W_{mask} \times (C_{mask} + 2)}$ is the combination of $\mat{F}_{mask}$ and a map of coordinates $\mat{O}_{x, y} \in \R^{H_{mask} \times W_{mask} \times 2}$. As described before, $\mat{O}_{x, y}$ is the relative coordinates from all the locations on $\mF_{mask}$ to $(x, y)$ (\ie, the location where the filters are generated). $MaskHead$ denotes the mask head, which consists of a stack of convolutions with dynamic parameters $\vec{\theta}_{x, y}$. $\mM^*_{x, y} \in \{0, 1\}^{H \times W \times C}$ is the mask of the instance associated with location $(x, y)$. $L_{dice}$ is the dice loss as in \cite{milletari2016v}, which is used to overcome the foreground-background sample imbalance. We do not employ focal loss here as it requires special initialization, which cannot be realized if the parameters are dynamically generated. Note that, in order to compute the loss between the predicted mask and the ground-truth mask $\mM^*_{x, y}$, they are required to have the same size. As mentioned before, the prediction is upsampled by $4$ and thus the resolution of the final prediction is half of that of the ground-truth mask $\mM^*_{x, y}$. We downsample $\mM^*_{x, y}$ by $2$ to make the sizes equal. These operations are omitted in Eq.~\eqref{loss_function_mask} for clarification.

Moreover, as shown in YOLACT \cite{bolya2019yolact}, the instance segmentation task can benefit from a joint semantic segmentation task. Thus, we also conduct experiments with the joint semantic segmentation task. However, unless explicitly specified, all the experiments in the paper are \emph{without} the semantic segmentation task. If used, the semantic segmentation loss is added to $L_{overall}$.

\subsection{Inference}
Given an input image, we forward it through the network to obtain the outputs including classification confidence $\vp_{x, y}$, center-ness scores, box prediction $\vt_{x, y}$ and the generated parameters $\vec{\theta}_{x, y}$. We first follow the steps in FCOS to obtain the box detections. Afterwards, box-based NMS with the threshold being $0.6$ is used to remove duplicated detections and then the top $100$ boxes are used to compute masks. Different from FCOS, these boxes are also associated with the filters generated by the controller. Let us assume that $K$ boxes remain after the NMS, and thus we have $K$ groups of the generated filters. The $K$ groups of filters are used to produce $K$ instance-specific mask heads. These instance-specific mask heads are applied, in the fashion of FCNs, to the $\mat{\tilde{F}}_{x, y}$ (\ie, the combination of $\mF_{mask}$ and $\mat{O}_{x, y}$) to predict the masks of the instances. Since the mask head is a very compact network (three $1 \times 1$ convolutions with $8$ channels and $169$ parameters in total), the overhead of computing masks is extremely small. For example, even with $100$ detections (\ie, the maximum number of detections per image on MS-COCO), only less $5$ milliseconds in total are spent on the mask heads, which only adds $\sim 10\%$ computational time to the base detector FCOS. In contrast, the mask head of Mask R-CNN has four $3 \times 3$ convolutions with $256$ channels, thus having more than 2.3M parameters and taking longer computational time.

\section{Experiments}
\setlength{\tabcolsep}{0.3pt}
\begin{table*}[t]
    \caption{Instance segmentation results with different architectures of the mask head on MS-COCO \texttt{val2017} split. ``depth": the number of layers in the mask head. ``width": the number of channels of these layers. ``time": the milliseconds that the mask head takes for processing $100$ instances.}
	\label{table:design_choice_mask_head}
	\begin{center}
		\subfloat[Varying the depth (width $= 8$). \label{varying_depth}]{
			\begin{tabular}{c|c|c|cc|ccc}
				\hline
				depth & time & AP & AP$_{50}$ & AP$_{75}$ & AP$_{S}$ & AP$_{M}$ & AP$_{L}$ \\
				\Xhline{2\arrayrulewidth}
				1 & \textbf{2.2} & 30.9 & 52.9 & 31.4 & 14.0 & 33.3 & 45.1 \\
				2 & 3.3 & 35.5 & 56.1 & 37.8 & 17.0 & 38.9 & 50.8 \\
				3 & 4.5 & \textbf{35.7} & \textbf{56.3} & 37.8 & 17.1 & \textbf{39.1} & 50.2  \\
				4 & 5.6 & \textbf{35.7} & 56.2 & \textbf{37.9} & \textbf{17.2} & 38.7 & \textbf{51.5} \\
				\hline
			\end{tabular}
		}
		\subfloat[Varying the width (depth $= 3$). \label{varying_width}]
		{
			\begin{tabular}{c|c|c|cc|ccc}
				\hline
				width & time & AP & AP$_{50}$ & AP$_{75}$ & AP$_{S}$ & AP$_{M}$ & AP$_{L}$ \\
				\Xhline{2\arrayrulewidth}
				2 & \textbf{2.5} & 34.1 & 55.4 & 35.8 & 15.9 & 37.2 & 49.1 \\
				4 & 2.6 & 35.6 & \textbf{56.5} & \textbf{38.1} & 17.0 & \textbf{39.2} & \textbf{51.4} \\
				8 & 4.5 & \textbf{35.7} & \textbf{56.3} & 37.8 & 17.1 & 39.1 & 50.2  \\
				16 & 4.7 & 35.6 & 56.2 & 37.9 & \textbf{17.2} & 38.8 & 50.8 \\
				\hline
			\end{tabular}
		}
	\end{center}
\end{table*}
\setlength{\tabcolsep}{1.4pt}

We evaluate \Ours\ on the large-scale benchmark MS-COCO \cite{lin2014microsoft}. Following the common practice \cite{he2017mask, tian2019fcos, lin2017focal}, our models are trained with split \texttt{train2017} (115K images) and all the ablation experiments are evaluated on split \texttt{val2017} (5K images). Our main results are reported on the \texttt{test}-\texttt{dev} split (20K images).

\subsection{Implementation Details}
Unless specified, we make use of the following implementation details. Following FCOS \cite{tian2019fcos}, ResNet-50 \cite{he2016deep} is used as our backbone network and the weights pre-trained on ImageNet \cite{deng2009imagenet} are used to initialize it. For the newly added layers, we initialize them as in \cite{tian2019fcos}. Our models are trained with stochastic gradient descent (SGD) over $8$ V100 GPUs for 90K iterations with the initial learning rate being $0.01$ and a mini-batch of $16$ images. The learning rate is reduced by a factor of $10$ at iteration $60K$ and $80K$, respectively. Weight decay and momentum are set as $0.0001$ and $0.9$, respectively. Following Detectron2 \cite{wu2019detectron2}, the input images are resized to have their shorter sides in $[640, 800]$ and their longer sides less or equal to $1333$ during training. Left-right flipping data augmentation is also used during training. When testing, we do not use any data augmentation and only the scale of the shorter side being $800$ is used. The inference time in this work is measured on a single V100 GPU with $1$ image per batch.

\setlength{\tabcolsep}{4pt}
\begin{table}[t]
	\caption{The instance segmentation results by varying the number of channels of the mask branch output (\ie, $C_{mask}$) on MS-COCO \texttt{val2017} split. The performance keeps almost the same if $C_{mask}$ is in a reasonable range, which suggests that \Ours\ is robust to the design choice.}
	\label{table:c_mask}
	\begin{center}
		\begin{tabular}{c|c|cc|ccc}
			\hline
			$C_{mask}$ & AP & AP$_{50}$ & AP$_{75}$ & AP$_{S}$ & AP$_{M}$ & AP$_{L}$ \\
			\Xhline{2\arrayrulewidth}
			1 & 34.8 & 55.9 & 36.9 & 16.7 & 38.0 & 50.1 \\
			2 & 35.4 & 56.2 & 37.6 & 16.9 & 38.9 & 50.4 \\
			4 & 35.5 & 56.2 & \textbf{37.9} & 17.0 & 39.0 & 50.8 \\
			8 & \textbf{35.7} & \textbf{56.3} & 37.8 & \textbf{17.1} & \textbf{39.1} & 50.2 \\
			16 & 35.5 & 56.1 & 37.7 & 16.4 & \textbf{39.1} & \textbf{51.2} \\
			\hline
		\end{tabular}
	\end{center}
\end{table}
\setlength{\tabcolsep}{1pt}

\setlength{\tabcolsep}{0.3pt}
\begin{table*}[h]
	\caption{Ablation study of the input to the mask head on MS-COCO \texttt{val2017} split. As shown in the table, without the relative coordinates, the performance drops significantly from $35.7\%$ to $31.4\%$ in mask AP. Using the absolute coordinates cannot improve the performance remarkably. If the mask head only takes as input the relative coordinates (\ie, no appearance in this case), \Ours\ also achieves modest performance.}
	\label{table:w_or_wo_offsets}
	\centering
		\begin{tabular}{c|c|c|c|cc|ccc|ccc}
			\hline
			w/ abs.\    coord.   & w/ rel.\  coord. & w/ ${\mF}_{mask}$ & AP & AP$_{50}$ & AP$_{75}$ & AP$_{S}$ & AP$_{M}$ & AP$_{L}$ & AR$_{1}$ & AR$_{10}$ & AR$_{100}$ \\
			\Xhline{2\arrayrulewidth}
			& & \checkmark & 31.4 & 53.5 & 32.1 & 15.6 & 34.4 & 44.7 & 28.4 & 44.1 & 46.2 \\
			& \checkmark & & 31.3 & 54.9 & 31.8 & 16.0 & 34.2 & 43.6 & 27.1 & 43.3 & 45.7 \\
			\checkmark & & \checkmark & 32.0 & 53.3 & 32.9 & 14.7 & 34.2 & 46.8 & 28.7 & 44.7 & 46.8 \\
			\hline
			& \checkmark & \checkmark & \textbf{35.7} & \textbf{56.3} & \textbf{37.8} & \textbf{17.1} & \textbf{39.1} & \textbf{50.2} & \textbf{30.4} & \textbf{48.8} & \textbf{51.5} \\
			\hline
		\end{tabular}
\end{table*}
\setlength{\tabcolsep}{1.4pt}

\subsection{Architectures of the Mask Head}
In this section, we discuss the design choices of the mask head in \Ours.
To our surprise, the performance is insensitive to the architectures of the mask head. Our baseline is the mask head of three $1 \times 1$ convolutions with $8$ channels (\ie, width $= 8$). As shown in Table~\ref{table:design_choice_mask_head} (3rd row), it achieves $35.7\%$ in mask AP. Next, we first conduct experiments by varying the depth of the mask head. As shown in Table~\ref{varying_depth}, apart from the mask head with depth being $1$, all other mask heads (\ie, depth $= 2, 3$ and $4$) attain similar performance. The mask head with depth being $1$ achieves inferior performance as in this case the mask head is actually a linear mapping, which has overly weak capacity. Moreover, as shown in Table~\ref{varying_width}, varying the width (\ie, the number of the channels) does not result in a remarkable performance change either as long as the width is in a reasonable range. We also note that our mask head is extremely light-weight as the filters in our mask head are dynamically generated. As shown in Table~\ref{table:design_choice_mask_head}, our baseline mask head only takes $4.5$ ms per $100$ instances (the maximum number of instances on MS-COCO), which suggests that our mask head only adds small computational overhead to the base detector. Moreover, our baseline mask head only has $169$ parameters in total. In sharp contrast, the mask head of Mask R-CNN \cite{he2017mask} has more than 2.3M parameters and takes $\sim 2.5 \times$ computational time ($11.4$ ms per $100$ instances).

\subsection{Design Choices of the Mask Branch}
We further investigate the impact of the mask branch. We first change $C_{mask}$, which is the number of channels of the mask branch's output feature maps (\ie, $\mF_{mask}$). As shown in Table~\ref{table:c_mask}, as long as $C_{mask}$ is in a reasonable range (\ie, from $2$ to $16$), the performance keeps almost the same. $C_{mask} = 8$ is optimal and thus we use $C_{mask} = 8$ in all other experiments by default.

As mentioned before, before taken as the input of the mask heads, the mask branch's output $\mF_{mask}$ is concatenated with a map of relative coordinates, which provides a strong cue for the mask prediction. As shown in Table~\ref{table:w_or_wo_offsets} (2nd row), the performance drops significantly if the relative coordinates are removed ($35.7\%$ vs. $31.4\%$). The significant performance drop implies that the generated filters not only encode the appearance cues but also encode the shape (and relative position) of the target instance. It can also be evidenced by the experiment only using the relative coordinates. As shown in Table~\ref{table:w_or_wo_offsets} (2rd row), only using the relative coordinates can also obtain decent performance ($31.3\%$ in mask AP). We would like to highlight that unlike Mask R-CNN, which encodes the shape of the target instance by a bounding-box, \Ours\ implicitly encodes the shape into the generated filters, which can easily represent any shapes including irregular ones and thus is much more flexible. We also experiment with the absolute coordinates, but it cannot largely boost the performance as shown in Table~\ref{table:w_or_wo_offsets} ($32.0\%$). This suggests that the generated filters mainly carry translation-invariant cues such as shapes and relative position, which is preferable.

\subsection{How Important to Upsample Mask Predictions?}
\setlength{\tabcolsep}{3pt}
\begin{table}[t]
	\caption{The instance segmentation results on MS-COCO \texttt{val2017} split by changing the factor used to upsample the mask predictions. ``resolution" denotes the resolution ratio of the mask prediction to the input image. Without the upsampling (\ie, factor $= 1$), the performance drops significantly. Almost the same results are obtained with ratio $2$ or $4$.}
	\label{table:upsampling}
	\begin{center}

		\begin{tabular}{c|c|c|cc|ccc}
			\hline
			factor & resolution & AP & AP$_{50}$ & AP$_{75}$ & AP$_{S}$ & AP$_{M}$ & AP$_{L}$ \\
			\Xhline{2\arrayrulewidth}
			$1$ & $1 / 8$ & 34.4 & 55.4 & 36.2 & 15.1 & 38.4 & 50.8 \\
			$2$ & $1 / 4$ & \textbf{35.8} & \textbf{56.4} & \textbf{38.0} & 17.0 & \textbf{39.3} & \textbf{51.1} \\
			$4$ & $1 / 2$ & 35.7 & 56.3 & 37.8 & \textbf{17.1} & 39.1 & 50.2 \\
			\hline
		\end{tabular}
	\end{center}
\end{table}
\setlength{\tabcolsep}{1.4pt}
As mentioned before, the original mask prediction is upsampled and the upsampling is of great importance to the final performance. We confirm this in the experiment. As shown in Table~\ref{table:upsampling}, without using the upsampling (1st row in the table), in this case \Ours\ can produce the mask prediction with $\frac{1}{8}$ of the input image resolution, which merely achieves $34.4\%$ in mask AP because most of the details (\eg, the boundary) are lost. If the mask prediction is upsampled by factor $= 2$, the performance can be significantly improved by $1.4\%$ in mask AP (from $34.4\%$ to $35.8\%$). In particular, the improvement on small objects is large (from $15.1\%$ to $17.0$), which suggests that the upsampling can greatly retain the details of objects. Increasing the upsampling factor to $4$ slightly worsens the performance (from $35.8\%$ to $35.7\%$ in mask AP), probably due to the relatively low-quality annotations of MS-COCO. We use factor $= 4$ in all other models as it has the potential to produce high-resolution instance masks.

\subsection{\Ours\ without Bounding-box Detection}
Although we still keep the bounding-box detection branch in \Ours, it is conceptually feasible to totally eliminate it if we make use of the NMS using no bounding-boxes. In this case, all the foreground samples (determined by the classification head) will be used to compute instance masks, and the duplicated masks will be removed by mask-based NMS. As shown in Table~\ref{table:mask_nms}, with the mask-based NMS, the same overall performance can be obtained as box-based NMS ($35.7\%$ vs. $35.7\%$ in mask AP).

\subsection{Comparisons with State-of-the-art Methods}
We compare \Ours\ against previous state-of-the-art methods on MS-COCO \texttt{test}-\texttt{dev} split. As shown in Table~\ref{table:comparisons_state_of_the_art_methods}, with $1\times$ learning rate schedule (\ie, $90K$ iterations), \Ours\ outperforms the original Mask R-CNN by $0.8\%$ ($35.4\%$ vs. $34.6\%$). \Ours\ also achieves a much faster speed than the original Mask R-CNN ($49$ms vs. $65$ms per image on a single V100 GPU). To our knowledge, it is the first time that a new and simpler instance segmentation method, without any bells and whistles outperforms Mask R-CNN both in accuracy and speed. \Ours\ also obtains better performance ($35.9\%$ vs.\  $35.5\%$) and on-par speed ($49$ms vs $49$ms) than the well-engineered Mask R-CNN in \texttt{Detectron2} (\ie, Mask R-CNN$^*$ in Table~\ref{table:comparisons_state_of_the_art_methods}). Furthermore, with a longer training schedule (\eg, $3\times$) or a stronger backbone (\eg, ResNet-101), a consistent improvement is achieved as well ($37.8\%$ vs. $37.5\%$ with ResNet-50 $3\times$ and $39.1\%$ vs. $38.8\%$ with ResNet-101 $3\times$). Moreover, as shown in Table~\ref{table:comparisons_state_of_the_art_methods}, with the auxiliary semantic segmentation task, the performance can be boosted from $37.8\%$ to $38.8\%$ (ResNet-50) or from $39.1\%$ to $40.1\%$ (ResNet-101), without increasing the inference time. For fair comparisons, all the inference time here is measured by ourselves on the same hardware with the official codes.

\setlength{\tabcolsep}{6pt}
\begin{table}[t]
	\caption{Instance segmentation results with different NMS algorithms. Mask-based NMS can obtain the same overall performance as box-based NMS, which suggests that \Ours\ can totally eliminate the box detection.}
	\label{table:mask_nms}
\centering
		\begin{tabular}{c|c|cc|ccc}
			\hline
			NMS & AP & AP$_{50}$ & AP$_{75}$ & AP$_{S}$ & AP$_{M}$ & AP$_{L}$ \\
			\Xhline{2\arrayrulewidth}
			box & \textbf{35.7} & 56.3 & \textbf{37.8} & 17.1 & 39.1 & 50.2 \\
			mask & \textbf{35.7} & \textbf{56.7} & 37.7 & \textbf{17.2} & \textbf{39.2} & \textbf{50.5} \\
			\hline
		\end{tabular}
\end{table}
\setlength{\tabcolsep}{1.4pt}

\setlength{\tabcolsep}{3pt}
\begin{table*}[h!]
	\caption{Comparisons with state-of-the-art methods on MS-COCO \texttt{test}-\texttt{dev}. ``Mask R-CNN" is the original Mask R-CNN \cite{he2017mask} and ``Mask R-CNN$^*$" is the improved Mask R-CNN in \texttt{Detectron2} \cite{wu2019detectron2}. ``aug.": using multi-scale data augmentation during training. ``sched.": the used learning rate schedule. 1$\times$ is $90K$ iterations, 2$\times$ is $180K$ iterations and so on. The learning rate is changed as in \cite{he2019rethinking}. `w/ sem": using the auxiliary semantic segmentation task.}
	\label{table:comparisons_state_of_the_art_methods}
	\begin{center}
	\begin{tabular}{l|c|c|c|c|cc|ccc}
		\hline
		method & backbone & aug. & sched. & AP & AP$_{50}$ & AP$_{75}$ & AP$_{S}$ & AP$_{M}$ & AP$_{L}$ \\
		\Xhline{2\arrayrulewidth}
		Mask R-CNN~\cite{he2017mask} & R-50-FPN & & $1\times$ & 34.6 & \textbf{56.5} & 36.6 & 15.4 & 36.3 & \textbf{49.7} \\
		\textbf{\Ours} & R-50-FPN & & $1\times$ & \textbf{35.4} & 56.4 & \textbf{37.6} & \textbf{18.4} & \textbf{37.9} & 46.9 \\
		\hline
		Mask R-CNN$^*$ & R-50-FPN & \checkmark & $1\times$ & 35.5 & 57.0 & 37.8 & 19.5 & 37.6 & 46.0 \\
		Mask R-CNN$^*$ & R-50-FPN & \checkmark & $3\times$ & 37.5 & 59.3 & 40.2 & 21.1 & 39.6 & 48.3 \\
		TensorMask~\cite{chen2019tensormask} & R-50-FPN & \checkmark & $6\times$ & 35.4 & 57.2 & 37.3 & 16.3 & 36.8 & 49.3 \\
		\textbf{\Ours} & R-50-FPN & \checkmark & $1\times$ & 35.9 & 56.9 & 38.3 & 19.1 & 38.6 & 46.8 \\
		\textbf{\Ours} & R-50-FPN & \checkmark & $3\times$ & 37.8 & 59.1 & 40.5 & 21.0 & 40.3 & 48.7 \\
		\textbf{\Ours} w/ sem. & R-50-FPN & \checkmark & $3\times$ & 38.8 & 60.4 & 41.5 & 21.1 & 41.1 & 51.0 \\
		\hline
		Mask R-CNN & R-101-FPN & \checkmark & $6\times$ & 38.3 & 61.2 & 40.8 & 18.2 & 40.6 & \textbf{54.1} \\
		Mask R-CNN$^*$ & R-101-FPN & \checkmark & $3\times$ & 38.8 & 60.9 & 41.9 & 21.8 & 41.4 & 50.5 \\
		YOLACT-700~\cite{bolya2019yolact} & R-101-FPN & \checkmark & $4.5\times$ & 31.2 & 50.6 & 32.8 & 12.1 & 33.3 & 47.1 \\
		TensorMask & R-101-FPN & \checkmark & $6\times$ & 37.1 & 59.3 & 39.4 & 17.4 & 39.1 & 51.6 \\
		\textbf{\Ours} & R-101-FPN & \checkmark & $3\times$ & 39.1 & 60.9 & 42.0 & 21.5 & 41.7 & 50.9 \\
		\textbf{\Ours} w/ sem. & R-101-FPN & \checkmark & $3\times$ & \textbf{40.1} & \textbf{62.1} & \textbf{43.1} & \textbf{21.8} & \textbf{42.7} & 52.6 \\
		\hline
	\end{tabular}
	\end{center}
\end{table*}

\begin{figure*}[t]
 	\centering
 	\includegraphics[width=.99999\linewidth]{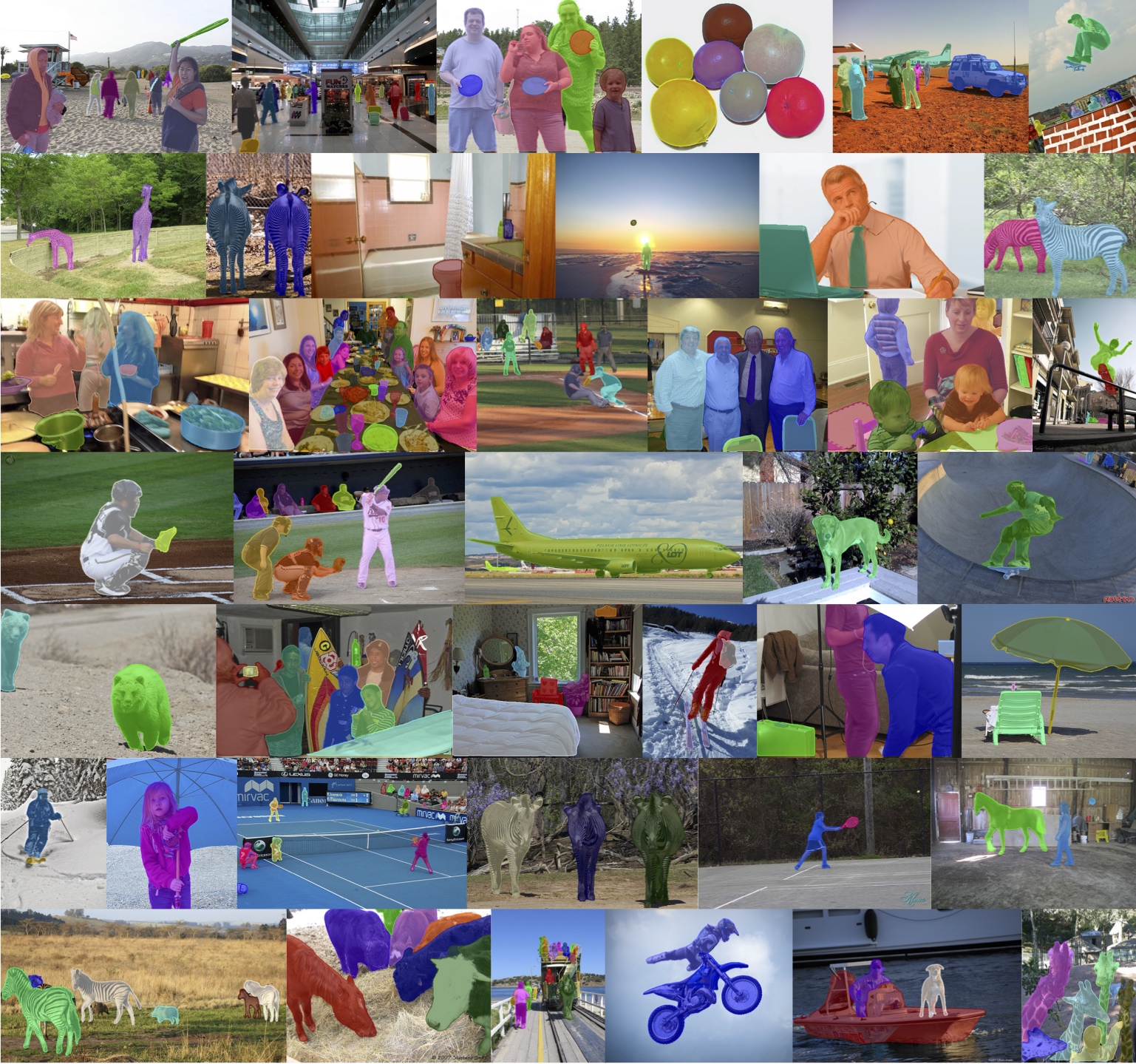}
 	\caption{More qualitative results of \Ours. Best viewed on screen.}
 	\label{fig:more_qualitative}
 \end{figure*}

We also compare \Ours\ with the recently-proposed instance segmentation methods. Only with half training iterations, \Ours\ surpasses TensorMask \cite{chen2019tensormask} by a large margin ($38.8\%$ vs. $35.4\%$ for ResNet-50 and $39.1\%$ vs. $37.1\%$ for ResNet-101). \Ours\ is also $\sim 8\times$ faster than TensorMask ($49$ms vs $380$ms per image on the same GPU) with similar performance ($37.8\%$ vs.\
$37.1\%$). Moreover, \Ours\ outperforms YOLACT-700 \cite{bolya2019yolact} by a large margin with the same backbone ResNet-101 ($40.1\%$ vs.\
$31.2\%$ and both with the auxiliary semantic segmentation task). Moreover, as shown in Fig.~\ref{fig:qualitative}, compared with YOLACT-700 and Mask R-CNN, \Ours\ can preserve more details and produce higher-quality instance segmentation results.
 More qualitative results are shown in Fig.~\ref{fig:more_qualitative}.

\section{Conclusions}
We have proposed a new and simpler instance segmentation framework, named \Ours. Unlike previous method such as Mask R-CNN, which employs the mask head with fixed weights, \Ours\ conditions the mask head on instances and dynamically generates the filters of the mask head. This not only reduces the parameters and computational complexity of the mask head, but also eliminates the ROI operations, resulting in a faster and simpler instance segmentation framework. To our knowledge, \Ours\ is the first framework that can outperform Mask R-CNN both in accuracy and speed, without longer training schedules needed. We believe that \Ours\ can be a new strong alternative to Mask R-CNN for instance segmentation.

\textbf{Acknowledgements}
CS was in part supported by ARC DP `Deep learning that scales'.

%
%
\bibliographystyle{splncs04}
\bibliography{1105}
\end{document}